\documentclass[10pt,twocolumn,letterpaper]{article}

\usepackage[pagenumbers]{iccv} 

\newcommand{\methodnameacr}{MID\xspace}
\newcommand{\methodName}{Mask, Inpaint and Denoise\xspace}

\definecolor{iccvblue}{rgb}{0.21,0.49,0.74}
\usepackage[pagebackref,breaklinks,colorlinks,allcolors=iccvblue]{hyperref}

\usepackage{algorithm}
\usepackage{algorithmic}
\usepackage{tabularx}
\usepackage{multirow}
\usepackage{colortbl}
\usepackage{xcolor}

\usepackage{amsthm}
\usepackage{enumitem}

\newtheorem{proposition}{Proposition}

\author{%
Hamadi Chihaoui \quad  Paolo Favaro \\
Computer Vision Group, University of Bern, Switzerland\\
\texttt{\{hamadi.chihaoui,paolo.favaro\}@unibe.ch}\\
}

\title{Unsupervised Real-World Denoising: Sparsity is All You Need}

\begin{document}
\maketitle

\begin{abstract}

Supervised training for real-world denoising presents challenges due to the difficulty of collecting large datasets of paired noisy and clean images. Recent methods have attempted to address this by utilizing unpaired datasets of clean and noisy images. Some approaches leverage such unpaired data to train denoisers in a supervised manner by generating synthetic clean-noisy pairs. However, these methods often fall short due to the distribution gap between synthetic and real noisy images.  
To mitigate this issue, we propose a solution based on input sparsification, specifically using random input masking. Our method, which we refer to as \methodName (\methodnameacr), trains a denoiser to simultaneously denoise and inpaint synthetic clean-noisy pairs. On one hand, input sparsification reduces the gap between synthetic and real noisy images. On the other hand, an inpainter trained in a supervised manner can still accurately reconstruct sparse inputs by predicting missing clean pixels using the remaining unmasked pixels.  
Our approach begins with a synthetic Gaussian noise sampler and iteratively refines it using a noise dataset derived from the denoiser's predictions. The noise dataset is created by subtracting predicted pseudo-clean images from real noisy images at each iteration. The core intuition is that improving the denoiser results in a more accurate noise dataset and, consequently, a better noise sampler. We validate our method through extensive experiments on real-world noisy image datasets, demonstrating competitive performance compared to existing unsupervised denoising methods.

\end{abstract}

\begin{figure}[t]
\centering\
   \setkeys{Gin}{width=0.45\linewidth}
    \captionsetup[subfigure]{skip=0.0ex,
                             belowskip=0.0ex,
                             labelformat=simple}
                             \setlength{\tabcolsep}{2.0pt}
    \renewcommand\thesubfigure{}
    \small

\small
  \begin{tabular}{cc}
 {\includegraphics{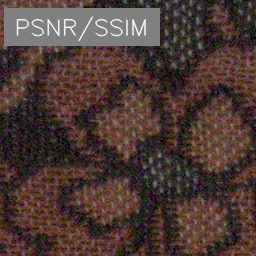}}& {\includegraphics{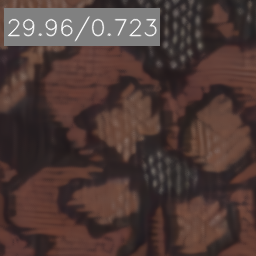}}\\
 (a)  \textit{ Noisy Input} & (b) \textit{ SDAP }\cite{pan2023random} \\
{\includegraphics{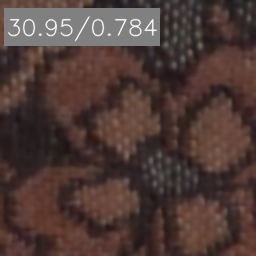}}& {\includegraphics{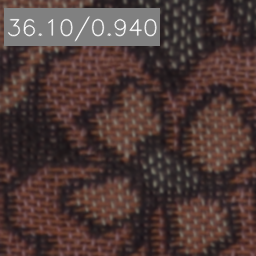}} \\
 
   (c)  \textit{ SCPGabNet }\cite{lin2023unsupervised} &  (d) \textit{ Ours (\methodnameacr)} \\
 \\
 
   \end{tabular}

\captionof{figure}{\label{fig:teaser}Visual comparison of unsupervised denoising methods on the SIDD validation dataset. Our method (\methodnameacr) preserves fine details better. Zoom in to see the reconstruction accuracy.}
\end{figure}

\section{Introduction}



Image denoising is one of the most long-studied problems in computer vision, thanks to its fundamental formulation that makes it the first step in testing image processing methods, as well as the realization that denoisers can serve multiple purposes \cite{elad2023imagedenoisingdeeplearning}. Traditional denoising methods rely on supervised learning, where pairs of clean and noisy images are available for training. However, in real-world scenarios, acquiring paired noisy-clean image datasets is often impractical. This has led to significant interest in unsupervised image denoising techniques, particularly in unpaired settings, where unrelated noisy and clean images (with no pairing) are available.

One intuitive approach is to use an Additive White Gaussian Noise (AWGN) sampler to generate a paired dataset of noisy-clean images and train a denoiser in a supervised manner. However, the mismatch between the synthetic noise distribution and real-world noise leads to poor performance at test time. In alternative, several unsupervised methods have also been proposed for the unpaired setting. These methods aim to learn or approximate the real noise distribution through adversarial training, typically involving a generative model and a discriminator. However, such methods often suffer from training instability and mode collapse, limiting their ability to effectively estimate real-world noise. This is due to the complexity and unknown nature of real-world noise distributions, often presenting local correlations. 

\begin{figure*}
    \centering
    \includegraphics[width=1\linewidth]{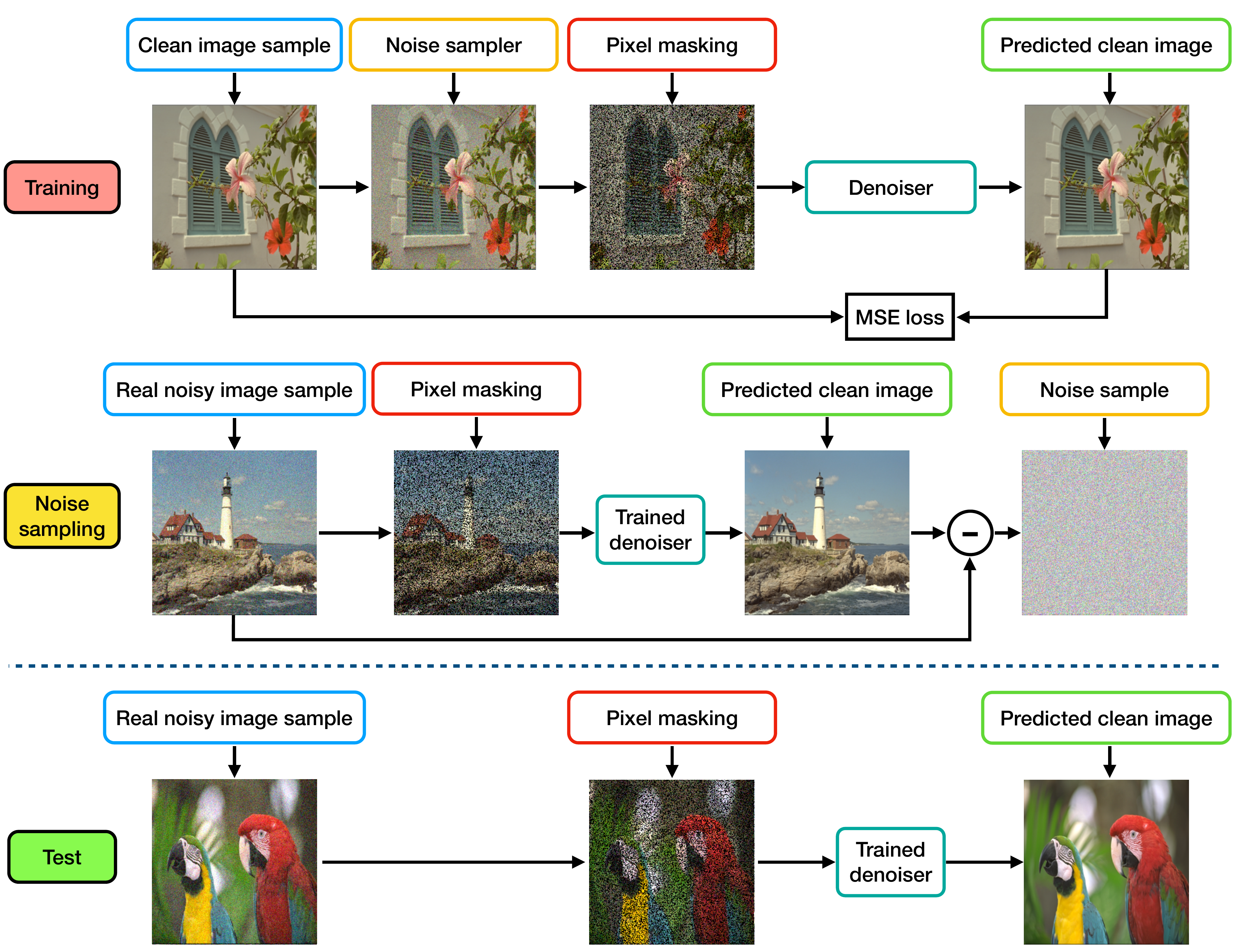}
    \caption{Overview of \methodnameacr. Top row: We show the processing steps used during supervised training. We use clean images from the available dataset, add synthetic noise (initially simply AWGN, and then later the noise samples are extracted from the real noisy images), mask the pixels and then train a denoiser to predict the clean image by minimizing a Mean Squared Error (MSE) loss. Middle row: To obtain better noise samples, we use the trained denoiser on the dataset of real noisy images. The noise samples are obtained simply by computing the residual between the predicted pseudo-clean image and the original noisy input. These residuals are then used as new noise samples in a new training of the denoiser. Bottom row: At test time we simply apply the trained denoiser on new real noisy images after applying masking. To further boost the accuracy, we average the predicted clean images for several random masks.}
    \label{fig:overview}
\end{figure*}

In this work, we explore a novel approach that \textit{removes the need for adversarial training} and bridges the gap between synthetic and real-world noise. We propose jointly training a denoiser to perform both denoising and inpainting on sparse inputs by using an AWGN sampler with random input masking to generate a synthetic paired dataset of noisy-clean images.
\textit{Our key idea is that randomly masking parts of both synthetic noisy images during training and real noisy images during testing reduces the distribution gap between training and testing phases.}
The missing content caused by masking can still be recovered by training the denoiser to inpaint the input simultaneously, leveraging the remarkable capabilities of deep neural networks and the inherent redundancy of natural images. Once the initial denoiser is trained, we propose a framework that iteratively refines the noise sampler by using the denoiser to generate a dataset of pseudo real-world noise. We show that by repeating this procedure iteratively, we can gradually improve the noise sampler and, more importantly, achieve a better-performing denoiser.
We incorporate these insights into a novel method called \methodName (\methodnameacr), which we elaborate on further in \cref{sec:mash} and illustrate in Figure~\ref{fig:overview}. In comparison to state-of-the-art unsupervised denoising methods, \methodnameacr
behaves favorably in terms of both quantitative metrics and
perceptual quality (see, for example, Figure~\ref{fig:teaser}).

Our contributions are summarized as follows
\begin{itemize}
 \item  We introduce \methodName (\methodnameacr), an innovative unsupervised image denoising method that utilizes random input masking to bridge the gap between the training phase (synthetic noise) and the testing phase (real noise). By randomly masking portions of the noisy image during both training and testing, we reduce the distribution mismatch between synthetic and real noise, enhancing the generalization ability of the denoiser to real-world noisy images without requiring paired noisy-clean datasets.
 
\item To the best of our knowledge, we are the \textbf{first} to apply random input masking in the context of unpaired image denoising, eliminating the need for adversarial training and its associated limitations.

    \item We propose an iterative procedure to refine noise samplers using residual noise from denoised real images. By progressively improving the sampler, our method enhances the denoiser’s performance and adaptability to real noise.
    \item \methodnameacr outperforms all unsupervised methods in the unpaired setting and is on par or better than the other unsupervised methods in real-world denoising across multiple datasets.
\end{itemize}


\section{Related Work}
\subsection{Non-learning-based image denoisers}
Traditional denoising algorithms, such as those found in  \cite{burger2012image, zhang2010two, talebi2013global}, define clean image properties using manually crafted priors. Some methods emphasize sparse representations \cite{bao20130sparsecoding, elad2006sparsecoding}, whereas others exploit the inherent recurrence of image patches \cite{zontak2013separating}. BM3D~\cite{dabov2009bm3d}, which employs collaborative filtering across similar image patches, is widely recognized for its strong performance on various benchmarks. Similarly, methods like NLM~\cite{buades2011nlm} and WNNM~\cite{gu2011nlm} also rely on leveraging related patches, using an implicit averaging strategy to reduce noise effectively.

\subsection{Supervised Image Denoising}
Supervised image denoising methods \cite{zamir2022restormer, zhang2017beyond, yue2019variational} train a neural network using a paired dataset of noisy images. \cite{zhang2017beyond} proposes a residual learning-based deep convolutional neural network (CNN) for image denoising, going beyond traditional Gaussian denoisers. By focusing on learning residual noise instead of directly predicting the clean image, the method effectively enhances denoising performance. \cite{yue2019variational} focuses on blind noise modeling and removal by leveraging a variational inference framework. It jointly learns noise distributions and denoises images, enabling the network to adapt to various types and levels of noise. Restormer \cite{zamir2022restormer} introduces an efficient transformer-based architecture specifically designed for high-resolution image restoration tasks. By leveraging multi-head self-attention mechanisms within a window-based framework, it achieves computational efficiency and superior performance. However, these methods rely on paired datasets, which may be expensive to acquire, thus limiting their applicability in practice.

\subsection{Unsupervised Image Denoising}
\noindent\textbf{Self-Supervised Image Denoising }
Self-supervised denoising methods train a denoiser using only a dataset of noisy images. Noise2Void \cite{krull2019noise2void} employs a masking strategy to split noisy images into input-target pairs. Blind-spot networks further enhance this approach by removing the corresponding noisy pixel from the input’s receptive field for each output pixel. To address information loss in the blind spot, probabilistic inference \cite{krull2020probabilistic, laine2019high} and regularization loss functions \cite{wang2022blind2unblind} have been introduced. However, these self-supervised approaches are often limited to noise that is spatially independent.
Some methods have also aimed to remove spatially correlated noise in a self-supervised manner. CVF-SID \cite{neshatavar2022cvf} separates noisy images into clean image and noise components. Among self-supervised approaches for real-world sRGB noise, AP-BSN \cite{lee2022ap} proposes asymmetric PD factors and post-refinement processing to better balance noise removal and aliasing artifacts, though it is computationally intensive during inference. SDAP \cite{pan2023random} generates random sub-samples from noisy images to create pseudo-clean targets, avoiding the need for paired clean-noisy datasets. LG-PBN \cite{wang2023lg} improves denoising by combining local and global patch-level features, using a blind-patch strategy to handle diverse noise patterns effectively in real-world images. Li et al. \cite{li2023spatially} introduce a spatially adaptive self-supervised learning method for real-world image denoising, where the model dynamically adjusts to different noise levels across image regions. AT-BSN \cite{chen2024exploring} employs efficient asymmetric blind-spots in self-supervised denoising to enhance performance in real-world scenarios.
However, those methods may be suboptimal from an information perspective, as they do not take advantage of the abundance of noise-free dataset \cite{deng2009imagenet, kuznetsova2020open, lin2014microsoft} available in the digital world.

\noindent\textbf{Unpaired Image Denoising }
Unpaired methods \cite{wu2020unpaired, jang2021c2n, lin2023unsupervised} address the challenge of data collection by training networks on datasets containing unpaired noisy and clean images. Many of these methods \cite{wu2020unpaired, jang2021c2n, lin2023unsupervised} focus on learning noise characteristics through adversarial training. This involves initially training a noise generator to replicate the noise distribution observed in noisy images, which is then used to transform clean images into synthetic noisy versions. The denoising network is subsequently trained using these synthetic noisy-clean image pairs. Additionally, Wu et al. \cite{wu2020unpaired} adopt a joint approach by training both a denoising network and a noise estimator simultaneously to model the noise distribution. The final denoising network is trained with a combination of synthetic noisy-clean pairs and noisy-denoised image pairs. Despite these efforts, accurately modeling noise distribution in the complex sRGB space remains challenging, limiting the effectiveness of unpaired methods for real-world photographs. SCPGabNet \cite{lin2023unsupervised} introduce an unsupervised approach to image denoising using two parallel generative adversarial branches that collaborate to reduce noise. These branches exchange and refine information, enabling the model to clean noisy images without requiring matched noisy-clean data pairs.
Our method falls within this category. Unlike previous works, \methodnameacr: 1) removes the need for adversarial training, 2) proposes a novel integration of masking with supervised training using synthetic noise, and 3) starts with a Gaussian noise sampler and iteratively refines it. To the best of our knowledge, these contributions have not been presented before.

\section{Unsupervised Image Denoising using \methodnameacr
\label{sec:mash}}

In this work, we assume access to a dataset of \textit{unpaired} noisy \(\mathcal{Y} = \{ y_1^r, \dots, y_n^r \}\) and clean images \(\mathcal{X} = \{ x_1, \dots, x_c \}\). 
In general, \(y^r\in\mathcal{Y}\) denotes one of the images corrupted with real noise, while \(x\in \mathcal{X}\) represents one of the clean images. In our approach $x$ and $y^r$ are not paired, that is, $y^r$ is a noisy image that has nothing to do with the clean image $x$. 
Our goal is to train a denoiser in a fully unsupervised manner by relying on these two datasets. In this section, we present our unsupervised denoising method \methodnameacr in detail.


\subsection{Limitations of Supervised Learning with Synthetic Data}

\begin{figure}
    \centering
    \includegraphics[width=0.8\linewidth]{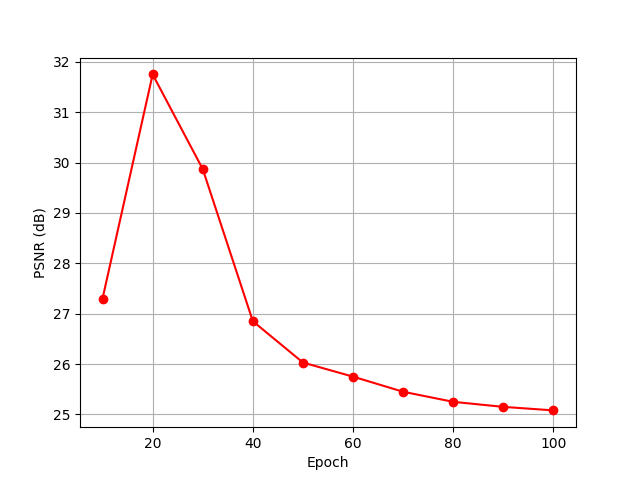}
    \caption{The denoiser performance on SIDD validation set when  trained with an AWGN sampler.} 
    \label{fig:awgn-denoising}
\end{figure}

One approach in the unpaired setting is to generate synthetic noisy-clean image pairs by adding synthetic noise to the clean images \(x\) and then by training a denoiser in a supervised manner. 
A simple choice to obtain noise samples is to use an Additive White Gaussian Noise (AWGN) sampler. To increase the computational efficiency, we draw a finite set of AWGN samples only once and then collect them in the set  $\mathcal{N}_0$. Then, at training time, whenever we need a new synthetic noise sample, we randomly select an element $n^s\sim \mathcal{N}_0$. Formally, given a clean image \(x\), the corresponding synthetic noisy image  is generated via \(y^s = x + n^s\). 
A denoiser $\mathcal{D}_0^s$ can then be trained by minimizing 
\begin{align}
\min_{\mathcal{D}_0^s} \sum_{x\in\mathcal{X},n^s\sim \mathcal{N}_0} \|\mathcal{D}_0^s(x+n^s) - x\|^2.
\end{align}
The main shortcoming of this approach is that such a denoiser tends to perform poorly at test time, when applied to real-world noisy images \(y^r\). 
This is due to the fact that during training the input is \(y^s = x + n^s\), whereas at test time, the input is \(y^r = x + n^r\). Unless real noise is AWGN, \(n^r\) and \(n^s\) will have different distributions and thus $y^r$ will be out of distribution for $\mathcal{D}_0^s$.
As shown in Figure~\ref{fig:awgn-denoising}, a denoiser  trained on clean images from the SIDD medium dataset with AWGN and applied to real noisy images from the SIDD validation dataset yields a poor performance. 


\subsection{Input Sparsification for Bridging the Synthetic-Real Distribution Gap \label{sect:3.2}}

To improve the performance of a denoiser trained on synthetic noisy-clean images, it is essential to minimize the distribution gap between the training (synthetic data) and testing (real data) phases. We propose a novel method to achieve this.
We propose reducing the train-test distribution gap by randomly masking the noisy inputs during both training and testing. 
We train the denoiser $\mathcal{D}_0$ by minimizing
\begin{align}
    \min_{\mathcal{D}_0} \sum_{\substack{x\in\mathcal{X}, n^s\sim\mathcal{N}_{0}\\ y^s = x+n^s, \mathbf{M}\sim\mathcal{M}_\alpha}}\| \mathcal{D}_0(\mathbf{M}\odot y^s) - x \|^2,
\end{align}
where \(\mathbf{M}\sim\mathcal{M}_\alpha\) is a binary random mask sample, $\mathcal{M}_\alpha$ is the set of all masks with the given masking ratio \(\alpha\) (in practice, we do not use a finite set, but randomly sample from a Bernoulli distribution at each pixel), and $\odot$ denotes the Hadamard element-wise (per pixel) product. This training involves simultaneously learning to denoise and to inpaint, and that is why we call our method \methodName, or in short, \methodnameacr. At test time, the denoiser is also applied to \(\mathbf{M} \odot y^r = \mathbf{M} \odot (x + n^r)\).


Input masking reduces the distribution gap between the synthetic training inputs and the real testing inputs. A formal proof is provided in Section \ref{masking-proposition} of the appendix. Intuitively, masking synthetic noisy images during training and real noisy images during testing increases the shared content between them (i.e., the masked pixels). In the extreme case where all pixels are masked (\(\mathbf{M} = \mathbf{0}\)), the two distributions become identical.

While input masking helps bridge the gap between synthetic and real-world noisy images, it can, however, negatively impact the training of the denoiser (which performs both denoising and inpainting simultaneously) by removing valuable input information. In fact, excessive masking may make the inpainting task overly ambiguous, leading to a poorly trained denoiser. This raises the question of how well an inpainter trained with randomly masked pixels can generalize to new images. In the next subsection, we address this aspect by examining the effect of the input masking ratio on the inpainter performance through experimental analysis.

\subsection{Achieving Accurate Reconstruction of Sparse Inputs with Supervised Inpainting \label{sect:3.3}}

We analyze the performance of an inpainter trained to reconstruct masked portions of an input image (\ie, the inpainter learns the mapping \(\mathbf{M} \odot x \rightarrow x\) -- \ie, no denoising) under different masking ratios between 20\% and 90\%. Since we are interested in the exact reconstruction of the original image (\ie, before masking), we evaluate the inpainted one based on how accurately the output image matches the original image. To achieve this, we train an  inpainter using the Imagenet validation dataset, where 90\% of the dataset is used for training and the remaining 10\% serves as a held-out test/validation set. The inpainter is trained to recover the masked pixels using the Mean Squared Error (MSE) loss via supervised learning.

\begin{figure}
    \centering
    \includegraphics[width=0.8\linewidth]{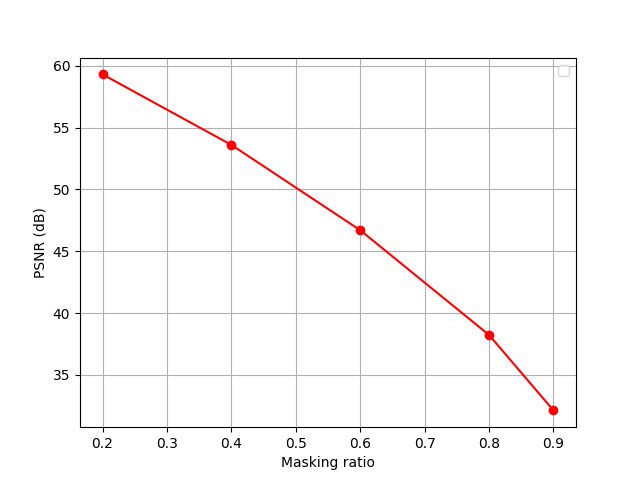}
    \caption{Inpainter performance on a validation set trained in a supervised way  at different random masking ratios. }
    \label{fig:inpainting_recoveribility}
\end{figure}

\begin{figure}[t]
\centering\
   \setkeys{Gin}{width=0.33\linewidth}
    \captionsetup[subfigure]{skip=0.0ex,
                             belowskip=0.0ex,
                             labelformat=simple}
                             \setlength{\tabcolsep}{0.0pt}
    \renewcommand\thesubfigure{}
    \small

\small
  \begin{tabular}{ccc}
 
 {\includegraphics{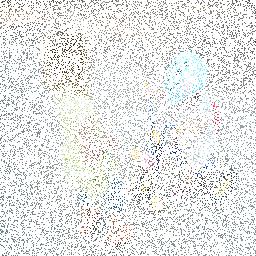}}&  {\includegraphics{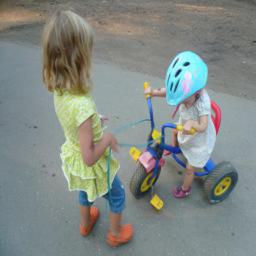}} &{\includegraphics{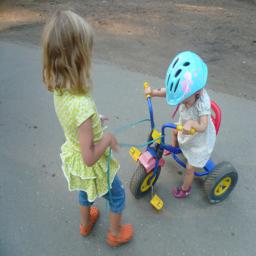}}\\
 \textit{Masked} & \textit{PSNR=39.87} & \textit{Original}  \\
{\includegraphics{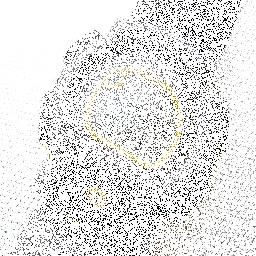}}& {\includegraphics{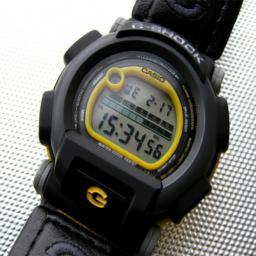}}& {\includegraphics{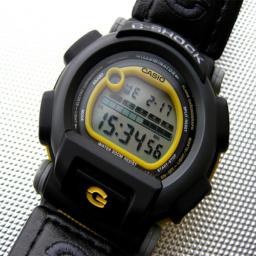}}\\
 \textit{Masked} &  \textit{PSNR=38.12}& \textit{Original}
 
   \end{tabular}

\captionof{figure}{\label{fig:inpainter-visual-results} Left: Masked input at 80\% of the pixels. Middle: The output of the inpainter on the image on the left. Right: Original images from the validation set.} 
\end{figure}

We assess the inpainter's performance on the held-out validation dataset during training.
Figure~\ref{fig:inpainting_recoveribility} illustrates the PSNR values of the inpainted images, comparing them to the original images on the validation set as a function of the masking ratio. We observe that the performance of the inpainter decreases with higher masking ratios. This is to be expected, as the task difficulty (and ambiguities) grow proportionally to the masking ratio. Nonetheless, even at an 80\% masking ratio a reconstruction PSNR above 35dB is excellent.
Figure \ref{fig:inpainter-visual-results} also shows the reconstruction of images when 80\% of the input pixels are masked. This surprisingly positive outcome can be explained by 1) the redundancy of information in natural images, which enables the inpainter to accurately predict masked pixels based on the observed ones, as well as 2) the advancement of Deep Neural Networks architectures and training. 

By combining (1) the findings from \cref{sect:3.2}, which show that random pixel masking reduces the gap between training and testing inputs, with (2) the experiments in Figure~\ref{fig:inpainting_recoveribility}, which demonstrate that a high masking ratio does not hinder the reconstruction of missing clean data due to the strong correlations in natural images, we suggest that a denoiser can be more effectively trained to jointly denoise and inpaint synthetic noisy-clean images while applying random input masking to the noisy data.  Once trained, the denoiser can be leveraged to obtain a better noise sampler by subtracting predicted pseudo-clean images from real noisy images. This process creates a virtuous cycle, where a denoiser trained in this manner benefits from synthetic noise samples that are more realistic (i.e., closer to real noise) than those used during its initial training. Consequently, a new denoiser trained on these improved samples can achieve even better denoising performance.



\subsection{Iterative Noise Sampler Boosting }
In this section, we assume to start with a well-trained denoiser (\(\mathcal{D}_0\) and explore the idea of iteratively refining a noise sampler to progressively better match the real noise distribution.

We obtain pseudo-real noise samples by applying the denoiser \(\mathcal{D}_0\) to the set of real noisy images \(y^r\in \mathcal{Y}\) after applying random masking. 
We compute the pseudo-real noise samples as the difference between the noisy image \(y^r\) and the predicted denoised images \(\mathcal{D}_0(\mathbf{M}\odot y^r)\), with $\mathbf{M}\sim\mathcal{M}_\alpha$. We define the set of pseudo-real noise samples as $\mathcal{N}_1\doteq \{y^r - \mathcal{D}_0(\mathbf{M}\odot y^r), \forall y^r\in\mathcal{Y} \text{ and } \mathbf{M}\sim\mathcal{M}_\alpha\}$.

If we had an accurate denoiser we would obtain a set of pseudo-real noise samples $\mathcal{N}_1$ that would fit well the distribution of real-world noise. Vice versa, if we used real-world noise as synthetic noise in the training of the denoiser, we would obtain a denoiser that would generalize very well on real noise images. \methodnameacr addresses this chicken and egg problem by using an iterative procedure that builds a positive gain over iteration time.
At iteration \(k\), \methodnameacr adds randomly selected pseudo-real noise samples from $n^p \in \mathcal{N}_k$ to the clean images $x$, randomly masks them with a binary mask $\mathbf{M}\sim\mathcal{M}_\alpha$, and then trains the denoiser  $\mathcal{D}_k$ to predict $x$ from $\mathbf{M}\odot (x+n^p)$ by solving
\begin{align}
    \min_{\mathcal{D}_k} \sum_{\substack{x\in\mathcal{X}, n^p\sim\mathcal{N}_{k},\\ \mathbf{M}\sim\mathcal{M}_\alpha}}\| \mathcal{D}_k(\mathbf{M}\odot (x+n^p)) - x \|.
    \label{eq:chicken}
\end{align}
The pseudo-real noise set $\mathcal{N}_k$ is then computed as
\begin{equation}
    \mathcal{N}_k  = \{ y^r - \mathcal{D}_{k-1}(\mathbf{M}\odot y^r), \forall y^r\in\mathcal{Y}, \mathbf{M}\sim\mathcal{M}_\alpha  \},
    \label{eq:noisesamples}
\end{equation}
except for the initial set $\mathcal{N}_0$, which consists of AWGN samples.
As shown in the experiments, although we start with AWGN samples, thanks to masking,  the denoiser is able to generalize well on real noisy images and to yield a set of pseudo-real noise samples $\mathcal{N}_1$ that is better than AWGN. We see experimentally that the noise samples $\mathcal{N}_k$ become more and more realistic over time, as the denoiser $\mathcal{D}_k$ improves its generalization performance over iteration time $k$.


\subsection{\methodName}

By putting all the previous steps together we can present our method for unsupervised image denoising in the unpaired data setting. We start with a cost-effective sampler \(\mathcal{N}_0\), \eg, an AWGN sampler, and use it to train a denoiser. The denoiser is then applied iteratively to generate a better noise sampler, as described in the previous section, which in turn improves the denoiser.
The steps for \methodnameacr are summarized in \cref{alg:d2d}.
At test time, given a real-world noisy image \(y^r\), the recovered clean image \(\hat{x}\) is obtained as an ensemble of \(K\) predictions
\begin{equation}
\hat{x} = \frac{1}{K} \sum_{p=1, \mathbf{M}\sim\mathcal{M}_\alpha}^K \mathcal{D}_m(\mathbf{M} \odot y^r),
\end{equation}
where for each prediction, a random binary mask \(\mathbf{M}\) is sampled and applied to the input image.

\begin{algorithm}[t!]
 \algsetup{linenosize=\normalsize}
  \normalsize
  \caption{\methodnameacr\label{alg:d2d}}
  \begin{algorithmic}[1]
  \REQUIRE Unpaired noisy \(\mathcal{Y}\) and clean \(\mathcal{X}\) datasets, total number of iterations $m$.
  \ENSURE Final denoising model $\mathcal{D}_m$\\
  \STATE Initialize the noise samples set $\mathcal{N}_0$ with AWGN samples.
  \FOR{$k  = 0 \textbf{\hspace{0.5pt} to \hspace{0.5pt}} m$}
   \STATE  Train a denoiser $\mathcal{D}_{k}$ by minimizing \cref{eq:chicken}
    
      \STATE  Build the noise sample set $\mathcal{N}_k$ as in \cref{eq:noisesamples}
  \ENDFOR
  \STATE Return denoising model $\mathcal{D}_m$\\
\end{algorithmic}

\end{algorithm}

\section{Experiments}
In this section, we will first introduce the experimental settings. We will then present the quantitative and qualitative results of \methodnameacr, along with comparisons with other methods.

\begin{table*}[h!]
  \begin{center}
    \renewcommand{\arraystretch}{1.2}
    \caption{\label{tab:real-results}Quantitative comparisons (PSNR(dB) and SSIM) of \methodnameacr and other real-world denoising methods on SIDD and DND datasets. The best results of the unsupervised approaches are marked in \textbf{bold}, while the second best ones are \underline{underlined}.}
     
    \scriptsize
    \begin{tabularx}{\linewidth}{
    @{\extracolsep{\fill}}llcccccc}
     

    \hline
    {Category} & {Method} & \multicolumn{2}{l}{SIDD Validation}& \multicolumn{2}{l}{SIDD Benchmark}& \multicolumn{2}{l}{DND Benchmark}\\
     \cline{3-8}
      &  &   PSNR $\uparrow$ & SSIM $\uparrow$ &  PSNR $\uparrow$ & SSIM $\uparrow$ &   PSNR $\uparrow$ & SSIM $\uparrow$ \\
     \hline
     
   \multirow{2}{*}{Non-learning based} & BM3D \cite{dabov2009bm3d}  & 25.71 & 0.576  & 25.65 & 0.685 & 34.51 & 0.851 \\
    & WNNM \cite{gu2014weighted} & 26.05 & 0.592  & 25.78 & 0.809 & 34.67 & 0.865 \\
    \hline
    \multirow{4}{*}{\shortstack[l]{Supervised \\ (Real pairs)}} & DnCNN \cite{zhang2017beyond} & 37.73 & 0.943  & 37.61 & 0.941 & 38.73 & 0.945 \\
    & Baseline, N2C \cite{ronneberger2015u} & 38.98 & 0.954  & 38.92 & 0.953 & 39.37 & 0.954 \\
    & VDN \cite{yue2019variational} & 39.29 & 0.956  & 39.26 & 0.955 & 39.38 & 0.952 \\
    & Restormer \cite{zamir2022restormer} & 39.93 & 0.960 & 40.02 & 0.960 & 40.03 & 0.956 \\
    \hline
    \multirow{6}{*}{\shortstack[l]{Self-Supervised}} & CVF-SID  \cite{neshatavar2022cvf} & 34.15 & 0.911 & 34.71 & 0.917 & 36.50 & 0.924\\
    & AP-BSN+R\textsuperscript{3} \cite{lee2022ap} & {36.74}  & {0.934} & {36.91}  & {0.931} & {38.09}  & {0.937} \\
    & Li et al. \cite{li2023spatially} & {37.39}  & {0.934} & \underline{37.41}  & {0.934} & {38.18}  & {0.938} \\
    & LG-BPN \cite{wang2023lg} & {-}  & {-} & {37.28}  & {0.936} & \textbf{38.43}  & \textbf{0.942}\\
      & SDAP \cite{pan2023random} & {37.30}  & {0.939} & {37.24/{37.79$^\dagger$ }}  & {0.936/0.886$^\dagger$ } & {37.86}  & {0.937}\\
       & AT-BSN \cite{chen2024exploring} & \underline{37.88}  & \underline{0.942} & \textbf{37.77}  & \textbf{0.946} & {38.29}  & \underline{0.939}\\
    \hline
     \multirow{4}{*}{Unpaired} &  Wu et al. \cite{wu2020unpaired} & - & - & - & - & 37.93 & {0.937}  \\
    & C2N \cite{jang2021c2n} & 35.36 & {0.932}  & 35.35 & \underline{0.937} & 37.28 & 0.924 
   \\
     & SCPGabNet \cite{lin2023unsupervised} & 36.53 & - & {36.53/36.98$^\dagger$} & {0.925/0.875$^\dagger$ } & 38.11 & \underline{0.939 } \\
      & Ours (\methodnameacr)  & \textbf{38.05} & \textbf{0.949} & \textbf{38.03}$^\dagger$  & \textbf{0.906}$^\dagger$  & \underline{38.38} & \textbf{0.942} \\
    
    \hline\\
    
    \end{tabularx}

  \end{center}

\end{table*}

\begin{figure*}[t]
    \setkeys{Gin}{width=0.19\linewidth}
    \captionsetup[subfigure]{skip=0.5ex,
                             belowskip=0.5ex,
                             labelformat=simple}
    \renewcommand\thesubfigure{}
    \setlength\tabcolsep{1.0pt}

\small
  \begin{tabular}{ccccc}
  \textit{Noisy input} & \textit{SDAP } \cite{pan2023random} & \textit{ SCPGabNet }\cite{lin2023unsupervised}  & \textit{Ours (\methodnameacr) } & \textit{Ground-truth}\\
{\includegraphics{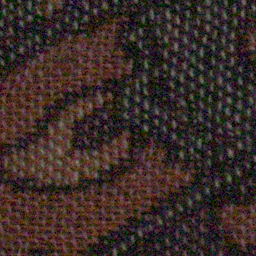}}& {\includegraphics{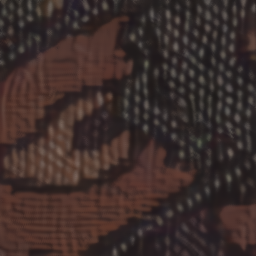}}& {\includegraphics{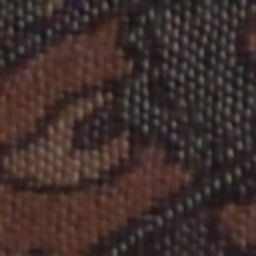}}& {\includegraphics{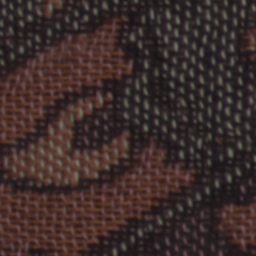}}& {\includegraphics{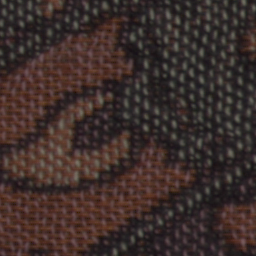}} \\
   & \textit{30.98/0.793} & \textit{31.99/0.841} & \textit{\textbf{35.96}/\textbf{0.944}} &  \\
  
 {\includegraphics{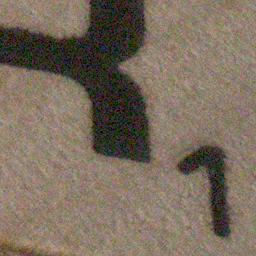}}& {\includegraphics{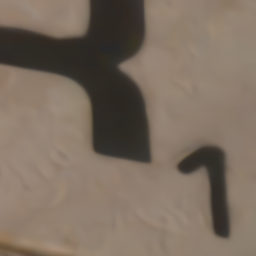}}& {\includegraphics{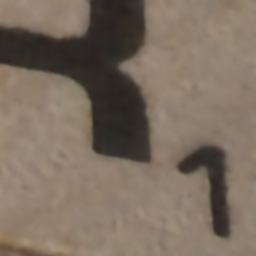}}& {\includegraphics{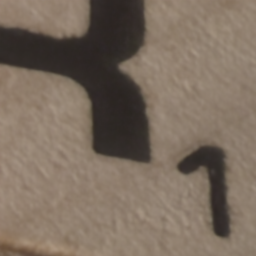}}& {\includegraphics{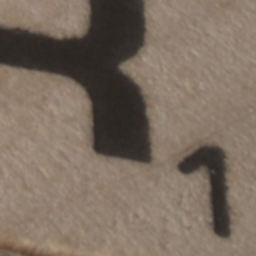}} \\
  & \textit{36.78/0.912} & \textit{37.80/0.931} & \textit{\textbf{39.00}/\textbf{0.952}} &  \\
{\includegraphics{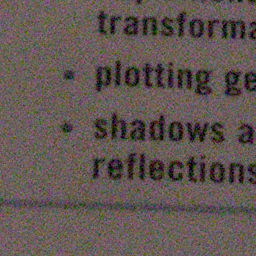}}& {\includegraphics{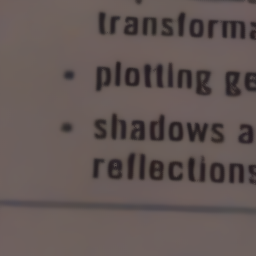}}& {\includegraphics{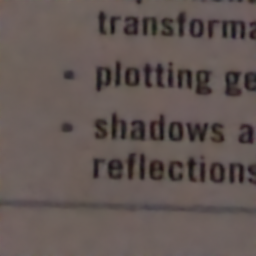}}& {\includegraphics{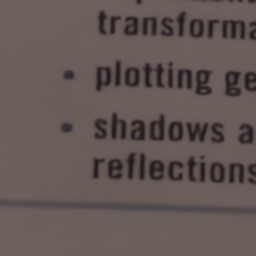}}& {\includegraphics{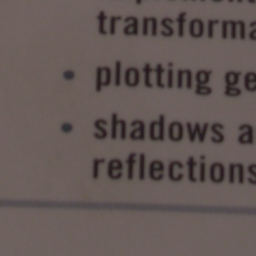}} \\
  & \textit{38.16/0.974} & \textit{37.87/0.972} & \textit{\textbf{40.12}/\textbf{0.979}} &  \\

\end{tabular}

\captionof{figure}{\label{fig:visual-results} Visual comparison on SIDD validation dataset. Zooming in is recommended to see the differences in reconstruction accuracy.}
\end{figure*}
\subsection{Experimental settings}
\noindent\textbf{Training and test data.} 
To train our method, we use the SIDD \cite{SIDD_2018_CVPR} Medium training set (which contains 320 pairs of noisy images and corresponding clean images captured using various smartphones). We begin by equally splitting the SIDD Medium training set into separate noisy and clean image groups. From this split, we use 160 clean images from one group and 160 noisy images from the other to create an unpaired dataset of real images for training the proposed algorithm. We evaluate our method using three widely recognized real-world noisy datasets: the SIDD~\cite{SIDD_2018_CVPR} Validation set, SIDD~\cite{SIDD_2018_CVPR} Benchmark, and the DND~\cite{plotz2017benchmarking} Benchmark. 
It is worth noting that the denoised images from the SIDD Benchmark and DND Benchmark can be uploaded to their respective websites to obtain PSNR and SSIM results. Unfortunately, the SIDD platform is currently unavailable and has been replaced by a Kaggle competition, which does not yet contain all the data. We reported our results and marked the entries with $^\dagger$. Additionally, we included some other unsupervised methods that have publicly shared their trained models to ensure a fairer comparison. It is also important to note that SSIM values may be calculated differently; the values reported by the Kaggle leaderboard reflect this variation. One explanation is that computing SSIM in the range [0, 1] versus [0, 255] can lead to significant differences in results.

\noindent\textbf{Implementation details.}
The network architecture for \methodnameacr follows the same structure as in NAFNet \cite{chen2022simple}. We train the denoiser for four rounds: the first round uses AWGN noise, while the remaining three rounds utilize residual noise derived from the last trained denoiser. For the masking ratio, we initially start at 80\% and then decrease it to 70\%. In the final round (when the denoiser is expected to perform better), we use a range of [50\%, 70\%]. It is important to note that the denoiser is retrained from scratch at each stage. The final denoised image is an ensemble of \(K=10\) predictions. The denoising network is trained from scratch for $2\cdot 10^{5}$ iterations using the Adam optimizer with initial learning rate of $4\cdot 10^{-4}$ and decayed to $ 10^{-6}$ using a cosine annealing scheduler.

\begin{table}[]
  
\centering
\caption{Effect of the Denoiser architecture for  \methodnameacr on the SIDD validation dataset. \label{tab:denoiser-architecture}}
     \centering
    \captionsetup{font=scriptsize}
\scriptsize
\begin{tabular*}{\linewidth}{
@{\extracolsep{\fill}}lcc@{}}
\toprule
Denoiser Architecture     & PSNR $\uparrow$ & SSIM $\uparrow$   \\ \midrule
DnCNN \cite{zhang2017beyond} & 36.85 & 0.934\\
NAFNet \cite{chen2022simple}   &      \textbf{38.05} & \textbf{ 0.949 }            \\

\bottomrule
\end{tabular*}
\end{table}

\begin{table}[]
  
\centering
\caption{Effect of the iterative noise sampler boosting method in  \methodnameacr on the SIDD validation dataset. \label{tab:denoiser-boosting}}
     \centering
    \captionsetup{font=scriptsize}
\scriptsize
\begin{tabular*}{\linewidth}{
@{\extracolsep{\fill}}lcccc@{}}
\toprule
Round  Number   & 1  & 2 & 4 & 6   \\ \midrule
PSNR & 36.38 & 37.43 & 38.05 & 37.96\\

\bottomrule
\end{tabular*}
\end{table}

\begin{table}[]
  
\centering
\caption{Effect of each component of  \methodnameacr on the SIDD validation dataset. \label{tab:analaysis}}
     \centering
    \captionsetup{font=scriptsize}
\scriptsize
\begin{tabular*}{\linewidth}{
@{\extracolsep{\fill}}lc}
\toprule
Method  & PSNR \\ \midrule
w/o masking   & 25.24    \\ 
w/o iterative refinement & 36.38 \\

w/o prediction ensembling & 37.68  \\
Ours & \textbf{38.05} \\

\bottomrule
\end{tabular*}
\end{table}

\begin{table}[]
  
\centering
\caption{Effect of the number $K$ of \methodnameacr predictions used in the ensemble on the SIDD validation dataset. \label{tab:denoiser-ensembling}}
     \centering
    \captionsetup{font=scriptsize}
\scriptsize
\begin{tabular*}{\linewidth}{
@{\extracolsep{\fill}}lcccc@{}}
\toprule
$K$   & 1  & 5 & 10 & 20  \\ \midrule
PSNR & 37.68 & 37.87 & 38.05 & 38.07\\

\bottomrule
\end{tabular*}
\end{table}

\begin{table}[]
  
\centering
\caption{Effect of the masking ratio in the first round of \methodnameacr  (AWGN denoising) on the SIDD validation dataset. \label{tab:masking-ratio}}
     \centering
    \captionsetup{font=scriptsize}
\scriptsize
\begin{tabular*}{\linewidth}{
@{\extracolsep{\fill}}lccccc@{}}
\toprule
Masking ratio   & 0\%  & 40\% & 60\% & 80\%  & 90\%  \\ \midrule
PSNR & 25.24 & 34.12 & 35.65 & \textbf{36.38} & 35.90\\

\bottomrule
\end{tabular*}
\end{table}

\noindent\textbf{Quantitative Comparison.}
Table 1 presents the quantitative comparison results for the SIDD  and DND  datasets. Our approach outperforms all existing unpaired methods   \cite{wu2020unpaired,jang2021c2n, lin2023unsupervised} by a significant margin of approximately 1.3 dB on the SIDD dataset. For self-supervised methods, \methodnameacr is on par or slightly better than the state-of-the-art AT-BSN \cite{chen2024exploring} and clearly outperforms all other self-supervised methods. When note that AT-BSN is a distillation-based method that trains multiple denoisers before distilling them while our method train a single network. When evaluated using PSNR and SSIM metrics, our method achieves performance levels comparable to DnCNN \cite{zhang2017beyond}, which has been trained on real-world paired datasets.

\noindent\textbf{Qualitative Comparison.}
The visual comparison of state-of-the-art self-supervised methods on the benchmark
datasets is shown in Figure \ref{fig:visual-results}.  \methodnameacr generates more detailed image compared to other methods.

\section{Ablation Study} 
We conduct extensive ablation studies on SIDD validation dataset to analyze the effectiveness of each component of \methodnameacr.


\subsection{Effect of each component of \methodnameacr}
In Table \ref{tab:analaysis}, we quantify the effect of the components of \methodnameacr: input masking, iterative refinement of the noise sampler, and prediction averaging. Masking is crucial, as training ends with overfitting without it. The iterative refinement of the noise sampler contributes around 1.7 dB to the final performance, while prediction ensembling contributes only 0.33 dB.

\subsection{Masking Ratio}
In Table \ref{tab:masking-ratio}, we present the effects of different masking ratios when training the denoiser with AWGN. A low masking ratio fails to bridge the gap between the training and testing distributions, leading to overfitting. 

\subsection{Iterative Noise Sampler Boosting }
In Table \ref{tab:denoiser-boosting},  we demonstrate the effect of our iterative boosting of the denoiser by reporting the PSNR of our denoiser after in each round. We observe that its performance gradually improves before eventually stagnating, with no further gains observed.

\subsection{Denoiser Network Architecture}

In Table \ref{tab:denoiser-architecture},  we present a comparison of the network architecture of our denoiser. With a more powerful network such as NAFNet \cite{chen2022simple}, we achieve better results. This suggests that our method could benefit from an improved network architecture.

\subsection{Prediction Ensemble Set Size $K$}

In Table \ref{tab:denoiser-ensembling}, we present the effect of the number of predictions used in our ensemble. Ensembling multiple predictions improves the denoising performance, and it almost saturates with $K=10$.


\section{Conclusions}

In this work, we propose a novel approach to solving real-world image unsupervised denoising eliminating the need for adversarial training and its associated limitations. We introduce a new method for integrating random masking to train a denoiser in a supervised manner, by allowing it to jointly denoise and inpaint images corrupted by noise. We iteratively improve our noise sampler by leveraging the denoiser's predictions and a dataset of real noise images. We demonstrate that \methodnameacr achieves state-of-the-art performance in unsupervised real-world image denoising.

{
    \small
    \bibliographystyle{ieeenat_fullname}
    \bibliography{main}
}

\clearpage
\appendix


\section{Impact of masking on the train-test distribution gap \label{masking-proposition}}

we show that masking reduces the distribution gap between the original training and testing inputs.

\begin{proposition}
\label{prop:gap}
Let $a^s = \mathbf{M}\odot y^s$ and $b^s = (1-\mathbf{M})\odot y^s$ be the visible and hidden pixels of a synthetic noisy image, and $a^r = \mathbf{M}\odot y^r$ and $b^r = (1-\mathbf{M})\odot y^r$ be the visible and hidden pixels of a real noisy image, respectively. Let us also denote with $p(a,b)$ the probability density function of synthetic data $y^s$ and with $q(a,b)$ the probability density function of real data $y^r$, where $a$ and $b$ are the visible and hidden entries of the images, respectively. Then, 
$D_\text{KL}(p(a,b) \| q(a,b)) \ge D_\text{KL}(p(a) \| q(a))$,
where $D_\text{KL}(p\|q)$ denotes the Kullback-Leibler divergence between $p$ and $q$.
\end{proposition}
\begin{proof}
    By using Bayes rule we have
    \begin{align}
    D_\text{KL}(p(a,b) \| q(a,b)) = \quad\quad &\nonumber\\
    \int p(a) D_\text{KL}(p(b|a) \| q(b|a)) da &+ D_\text{KL}(p(a) \| q(a)) =\nonumber\\
    &\ge D_\text{KL}(p(a) \| q(a)),\nonumber
\end{align}
since $p(a)\ge 0$ and also $D_\text{KL}(p(b|a) \| q(b|a))\ge 0$.
\end{proof}
\section{Additional qualitative comparison}

We provide an additional qualitative comparison on the SIDD validation dataset in Figure~\ref{fig:visual-results}. We can see visually that \methodnameacr tends to output restored images with texture at a higher level of detail and while reducing noise more than other unsupervised methods.

\section{Efficiency comparison}

We provide an efficiency comparison in Table \ref{tab:additional_comparisons}. For a fair comparison with AT-BSN, which is a distillation method, we distill our model using the network they used and denote this method as Ours(distilled).  Our vanilla method maintains reasonable efficiency, while our distilled version outperforms AT-BSN.

\begin{table}[H]
\setlength{\tabcolsep}{4pt}
\caption{Performance/efficiency trade-off on SIDD validation.\label{tab:additional_comparisons}}
     \centering
    \footnotesize
\vspace{-0.3cm}
\begin{tabular}{@{}lccccc@{}}
\toprule
   Method  & SCPGabNet  & AT-BSN  & Ours(K=10) & Ours(distilled) \\ \midrule
 Time (ms)              &     \underline{62}     & \textbf{32}         & 201   & \textbf{32}      \\
PSNR (dB)            &  36.53    & 37.88        & \textbf{38.05}     & \underline{37.96}   \\ 
\bottomrule
\end{tabular}
\vspace{-.6cm}   
\end{table}

\begin{figure*}[t]
    \setkeys{Gin}{width=0.2\linewidth}
    \captionsetup[subfigure]{skip=0.5ex,
                             belowskip=0.5ex,
                             labelformat=simple}
    \renewcommand\thesubfigure{}
    \setlength\tabcolsep{1.0pt}

\small
  \begin{tabular}{ccccc}
  \textit{Noisy input} & \textit{SDAP } \cite{pan2023random} & \textit{ SCPGabNet }\cite{lin2023unsupervised}  & \textit{Ours } & \textit{Ground-truth}\\
{\includegraphics{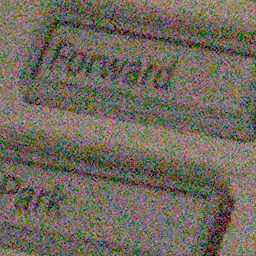}}& {\includegraphics{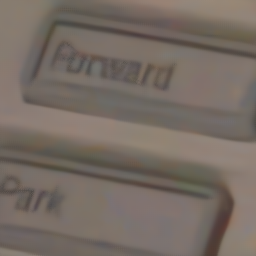}}& {\includegraphics{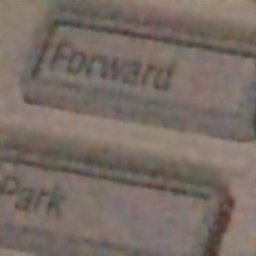}}& {\includegraphics{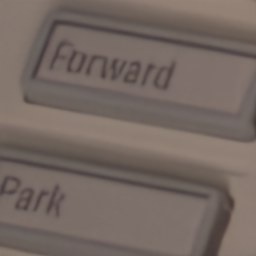}}& {\includegraphics{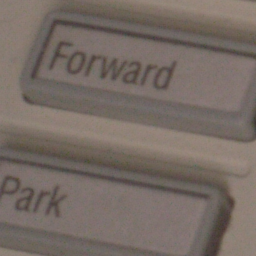}} \\

{\includegraphics{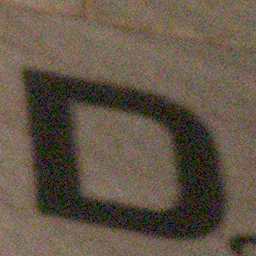}}& {\includegraphics{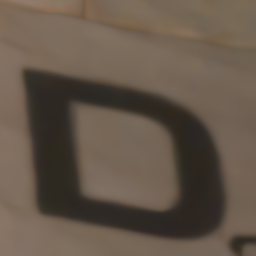}}& {\includegraphics{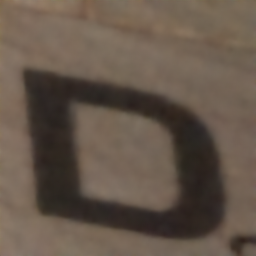}}& {\includegraphics{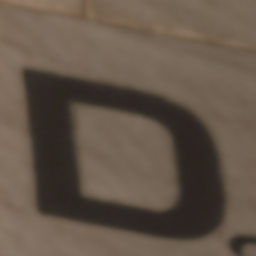}}& {\includegraphics{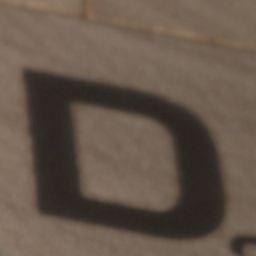}} \\
  
{\includegraphics{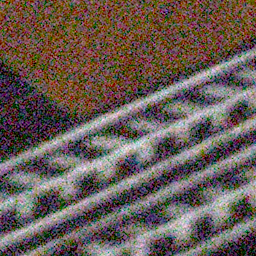}}& {\includegraphics{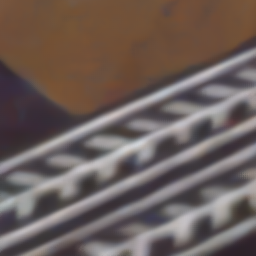}}& {\includegraphics{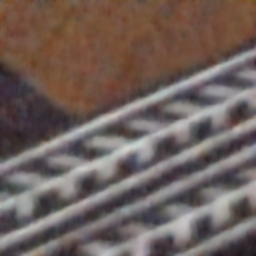}}& {\includegraphics{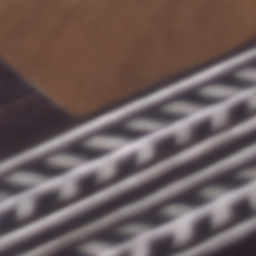}}& {\includegraphics{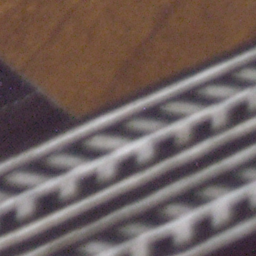}} \\

{\includegraphics{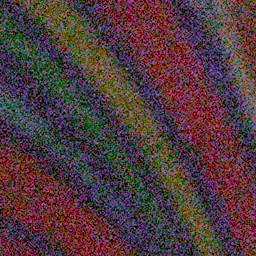}}& {\includegraphics{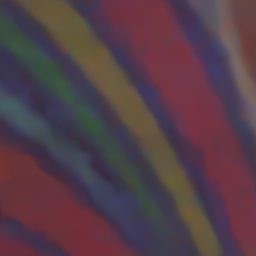}}& {\includegraphics{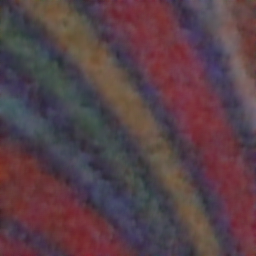}}& {\includegraphics{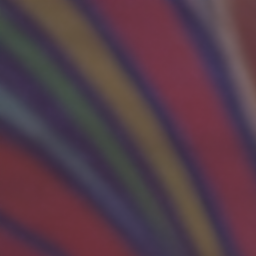}}& {\includegraphics{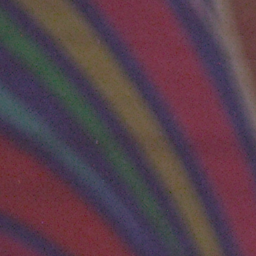}} \\

  {\includegraphics{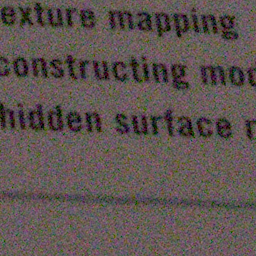}}& {\includegraphics{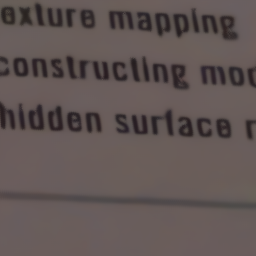}}& {\includegraphics{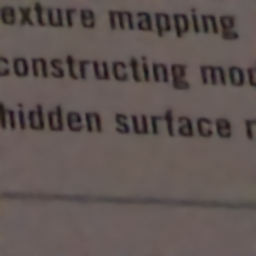}}& {\includegraphics{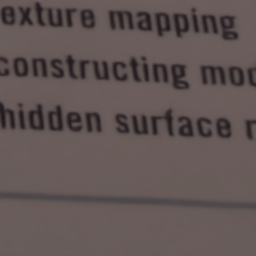}}& {\includegraphics{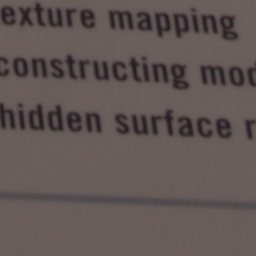}} \\

\end{tabular}

\captionof{figure}{\label{fig:visual-results} Visual comparison on SIDD validation dataset. Zooming in is recommended to see the differences in reconstruction accuracy.}
\end{figure*}

\end{document}